\documentclass{article}
\usepackage{acra}
\usepackage{graphicx}
\usepackage{url} 
\usepackage{float}
\usepackage{subcaption}
\usepackage{amsmath}
\usepackage{pifont}

\newcommand{\cmark}{\ding{51}} 
\newcommand{\xmark}{\ding{55}} 

\title{Performance Assessment of Lidar Odometry Frameworks: A Case Study at the Australian Botanic Garden Mount Annan}
\author{Mohamed Mourad Ouazghire$^1$, Julie Stephany Berrio$^2$, Mao Shan$^2$, Stewart Worrall$^2$ \\ $^1$ Department of Computer Science, Technical University of Darmstadt, Germany \\ $^2$ The Australian Center For Robotics, The University of Sydney, Australia \thanks{This work was supported by University of Sydney (Australian Centre for Robotics), Constructor University Bremen, Technical University of Darmstadt, with a funding grant from the German Academic Exchange Service (DAAD). } \\ 
mourad.ouazghire@stud.tu-darmstadt.de, \{stephany.berrioperez, mao.shan, stewart.worrall\}@sydney.edu.au}

\begin{document}

\maketitle

\begin{abstract}
Autonomous vehicles are being tested in diverse environments around the world. However, there is a notable gap in the evaluation of datasets representing natural unstructured environments such as forests or gardens. To address this, we present a study on localization at the Mount Annan Botanic Garden. This area encompasses open grassy areas, paved pathways, and densely vegetated sections with trees and other objects. The data set was recorded using a 128-beam LiDAR sensor and GPS and IMU readings to track the ego vehicle.
This paper evaluates the performance of two state-of-the-art LiDAR-inertial odometry frameworks, COIN-LIO and LIO-SAM, on this dataset. We analyze trajectory estimates in both horizontal and vertical dimensions and assess relative translation and yaw errors over varying distances. Our findings reveal that while both frameworks perform adequately in the vertical plane, COIN-LIO demonstrates superior accuracy in the horizontal plane, particularly over extended trajectories. In contrast, LIO-SAM shows an increase in drift and yaw errors over longer distances.

\end{abstract}

\section{Introduction}

Combining LiDARs with inertial measurement units (IMUs) can achieve higher accuracy and robustness by fusing complementary sensor data. However, these systems often face challenges related to cumulative drift over time, especially when traversing long distances. 

\begin{figure}[t]
    \centering
    \includegraphics[width=7cm]{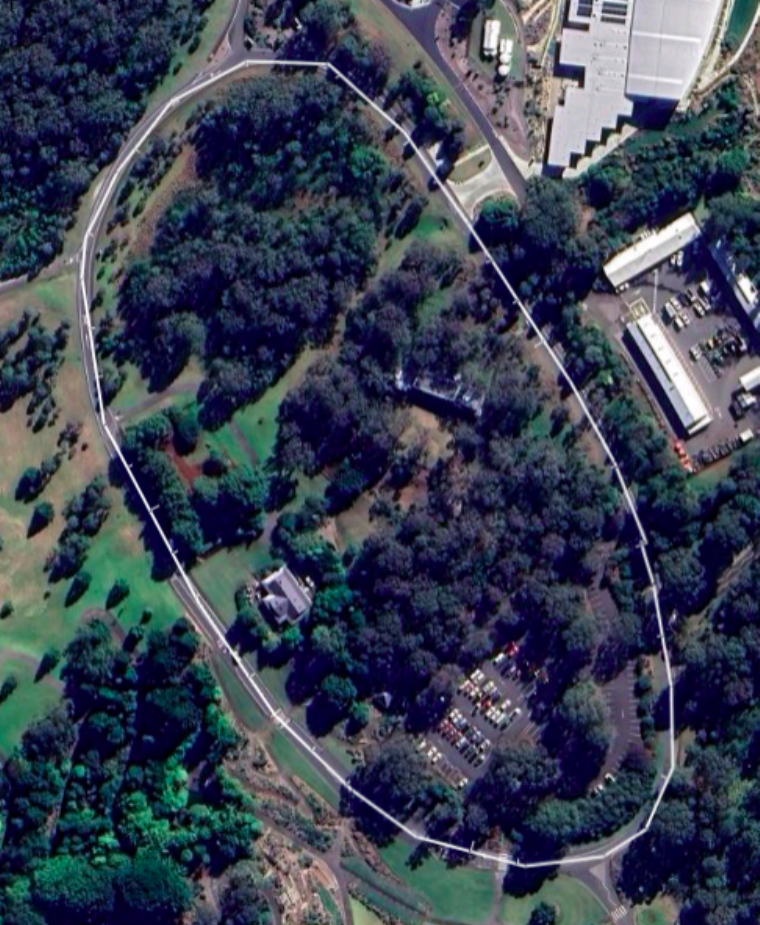}
    \caption{Australian Botanic Garden Mount Annan 3D Satellite Image}
\end{figure}

COIN-LIO[1] and LIO-SAM[2] are two prominent LiDAR-inertial odometry frameworks that have gained attention for their utility in two different scenarios. COIN-LIO is known to improve the robustness of LiDAR-inertial odometry in geometrically degenerate scenarios, like tunnels or flat fields. LIO-SAM, on the other hand, is known for its adaptability in Mixed Indoor-Outdoor and urban environments. 

Despite advances in these frameworks, there remains a need for comprehensive evaluations in diverse and challenging environments. The Mount Annan National Botanic Garden in Fig. 1 provides an ideal setting for such an evaluation due to its complex terrain and rich vegetation, which pose significant challenges for SLAM systems. The presence of a loop path and varying environmental conditions make it suitable for testing the robustness and accuracy of loop closure mechanisms. 

Other public datasets such as WildPlaces[3], is intended for use in lidar place recognition in unstructured natural environments, and is collected with a handheld sensor payload. MulRan[4] is another large-scale data set that captures mixed urban and suburban sequences taken with vehicle mounted systems in various terrains.

The Mount Annan Australian Botanic Garden, depicted in Fig. 1, presents an ideal setting in its turn due to its looped path and densely vegetated roadways. Through this vehicle-mounted system, we are complementing gaps that current datasets do not address.

This paper aims to evaluate and compare the performance of COIN-LIO and LIO-SAM when applied to the collected dataset.

\begin{figure*}[!t]
\centering
\renewcommand{\arraystretch}{1.8} 
\setlength{\tabcolsep}{4pt} 
\resizebox{\textwidth}{!}{%
\begin{tabular}{|l|c|c|c|c|c|}
\hline
\textbf{Criteria}                        & \textbf{LIO-SAM} & \textbf{COIN-LIO}          & \textbf{LEGO-LOAM}         & \textbf{Imaging Lidar Place Localization} & \textbf{FAST-LIO} \\ \hline
\textbf{Real-time Performance}           & \cmark           & \cmark                     & \cmark                     & Variable                                   & \cmark            \\ \hline
\textbf{Tightly-coupled IMU Integration} & \cmark           & Scan-Context Compatible    & \xmark                     & \xmark                                     & \cmark            \\ \hline
\textbf{Loop Closure}                    & \cmark           & Scan-Context Compatible    & Scan-Context Compatible    & \cmark                                     & Scan-Context Compatible \\ \hline
\textbf{Computational Efficiency}        & Moderate         & High                       & High                       & Moderate                                   & High             \\ \hline
\textbf{Mapping Capability}              & \cmark           & \cmark                     & Low                        & \xmark                                     & Limited           \\ \hline
\textbf{Robustness to Dynamic Environments} & Moderate      & High                       & Low                        & High                                      & Moderate          \\ \hline
\end{tabular}%
}
\caption{Comparison table of Lidar Inertial Odometry Methods}
\label{tab:lidar_methods}
\end{figure*}

\section{Related Work}

\subsection{LiDAR-Inertial Odometry}
LiDAR-inertial odometry combines LiDAR data with inertial measurements to provide continuous and accurate localization. The IMU provides high-frequency motion estimates, helping bridge gaps between LiDAR scans, especially in dynamic or geometrically challenging environments, such as tunnels, flat fields, or dense forests. Several advanced frameworks have been developed to enhance this synergy and improve the robustness of the system, particularly with Ouster sensors.

\textbf{LEGO-LOAM}[5] is a foundational framework that optimizes the structure of traditional LOAM by reducing computational complexity while maintaining real-time performance. However, it does not natively integrate IMU data, which can limit its performance in high-speed dynamic environments.

\textbf{LIO-SAM}, on the other hand, directly addresses this by tightly coupling LiDAR and IMU data through a factor graph approach, making it well suited for aggressive motion scenarios and reducing drift. This integration significantly improves its performance with Ouster sensors in particular.

Moreover, \textbf{Imaging Lidar Place Recognition}[6] focuses on robust place recognition and loop closure, which is essential to ensure global consistency in long-duration SLAM systems. This framework complements odometry methods and performs well with multibeam imaging lidars, enhancing the overall robustness in challenging conditions. 

Lastly, a notable recent development is \textbf{COIN-LIO}, an innovative framework introduced in 2024. Built on the efficient \textit{FAST-LIO2}[7] architecture, it improves the robustness of LiDAR-inertial odometry in geometrically degenerate environments such as tunnels and flat fields. By projecting LiDAR intensity returns into an image and introducing a novel image processing pipeline, it produces filtered images with enhanced brightness consistency(Fig. 3), especially while being highly compatible with Ouster sensors. COIN-LIO further leverages intensity as an additional modality by employing a new feature selection scheme that detects uninformative directions in point cloud registration and explicitly selects patches with complementary image information. Photometric error minimization in these image patches is then fused with inertial measurements and point-to-plane registration in an iterated Extended Kalman Filter.

\begin{figure}[H]
    \centering
    \includegraphics[width=8cm, height=5cm]{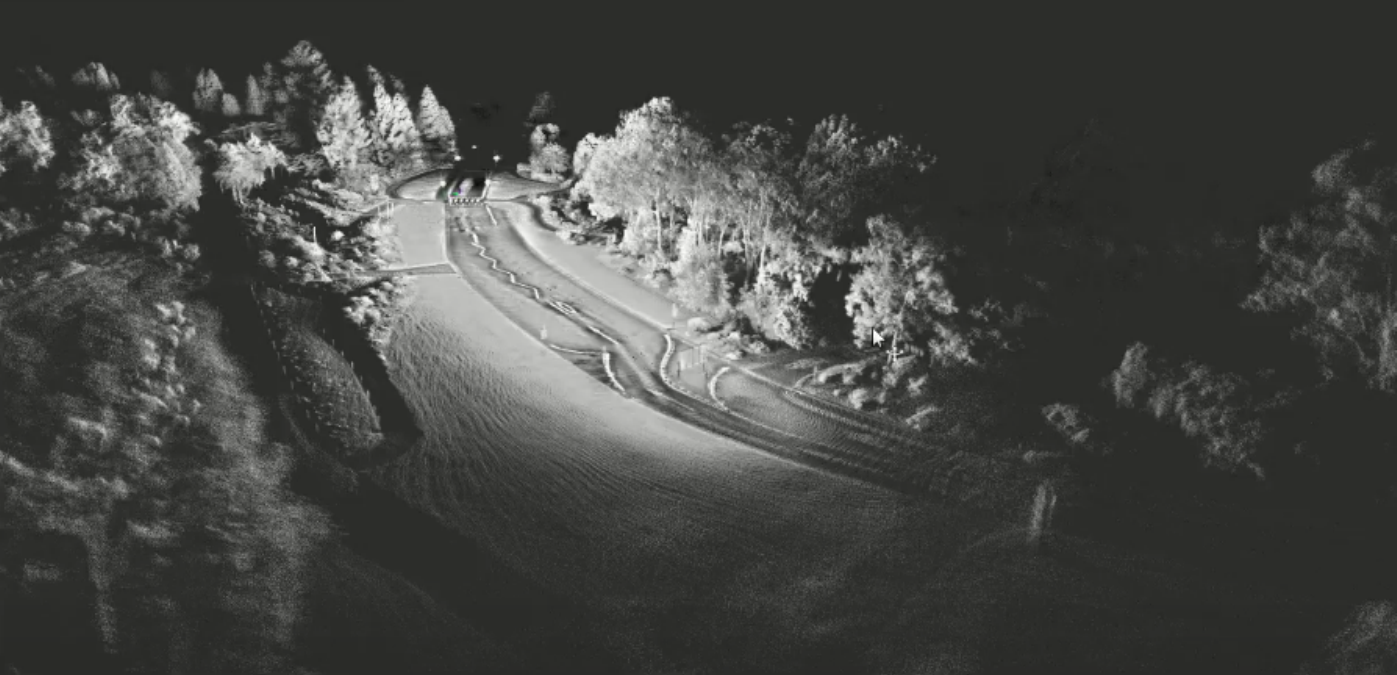}
    \includegraphics[width=8cm, height=2cm]{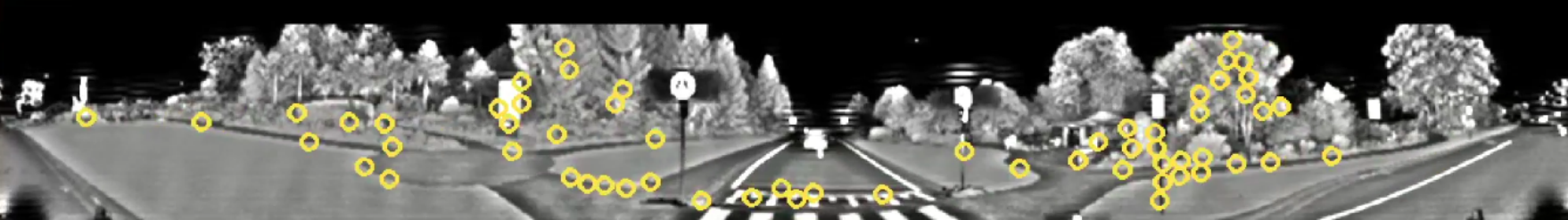}
    \caption{Body frame's image projection visualized through Coin-LIO at the Australian Botanic Garden Mount Annan}
\end{figure}

We set a comparative table \textit{(Fig. 2)} to distinguish between the characteristics of each of the frameworks mentioned above, prior to starting the experimental setup. 

Considering the excellent compatibility of LIO-SAM and COIN-LIO with Ouster sensors, and their superiority over the compared frameworks in the above criteria, we decided to conduct the trajectory evaluation of the dataset on these promising frameworks.

\subsection{Trajectory Evaluation Tools}

Evaluating the performance of LiDAR-inertial odometry systems requires precise tools for comparing estimated trajectories against ground-truth data. Two prominent tools in this domain are the \textbf{RPG Trajectory Evaluation Toolbox}[8] and \textbf{EVO} (Evolution)[9].

The \textbf{RPG Trajectory Evaluation Toolbox}, developed by the Robotics and Perception Group at the University of Zurich, offers a comprehensive framework for trajectory analysis. Provide detailed metrics such as the Absolute Trajectory Error (ATE) and Relative Pose Error (RPE), which are essential for assessing both global consistency and local accuracy of the trajectories. The toolbox is known for its robustness and ease of integration with various datasets and systems, making it a preferred choice for researchers who focus on high-precision evaluations.

\textbf{EVO} is another widely used tool to evaluate odometry and SLAM algorithms. Supports multiple data formats and offers visualization capabilities in addition to quantitative metrics. EVO is particularly user-friendly and compatible with common data sets such as TUM RGB-D [10] and KITTI [11], allowing flexible application in different evaluation scenarios.

In addition to these, other tools such as \textit{tum-eval} and \textit{KITTI odometry evaluation toolkit} are more specialized for their respective datasets.

In the context of this study, the \textbf{RPG Trajectory Evaluation Toolbox} was selected over \textbf{EVO} due to its advanced analytical capabilities and better alignment with the evaluation criteria of our LiDAR-inertial odometry systems. The detailed error metrics in the RPG toolbox and seamless integration with our experimental setup provided more insightful results, facilitating a deeper understanding of the system's performance.

\subsection{Loop Closure Techniques}

In LiDAR-based systems, loop closure often involves matching current scans with previously stored scans or submaps. \textit{Fig. 4} illustrates a top-side view of a point cloud looped trajectory generated in the context of this study. Iterative Closest Point[12] and Normal DistributionsTransform[13] are common algorithms used for scan matching. Feature-based methods extract key features from the point clouds to facilitate matching.

Recent approaches have used learning-based methods to improve loop closure detection. For example, SegMatch uses segment-based place recognition to identify loop closures in 3D point clouds [9]. Others have incorporated semantic information to enhance matching robustness in dynamic environments.

\begin{figure}[h]
    \centering
    \includegraphics[width=0.5\textwidth]{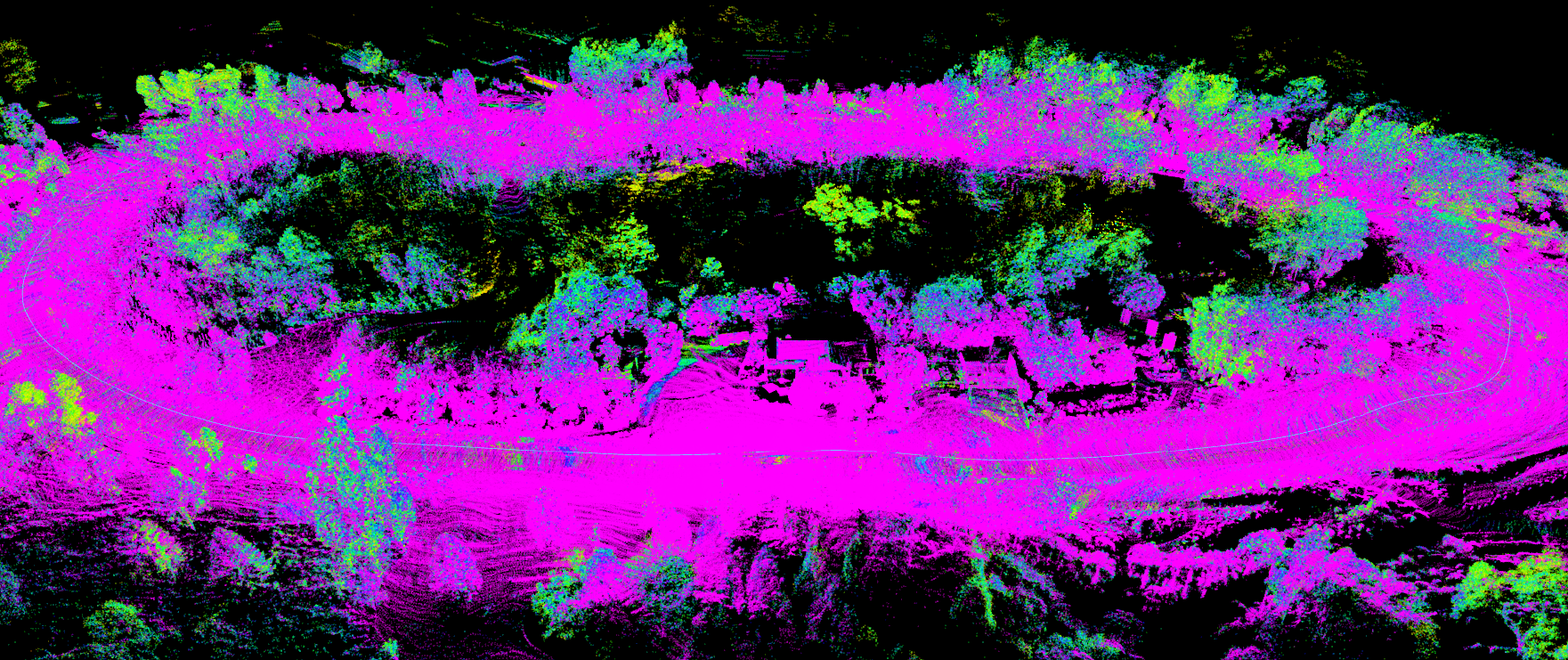}
    \caption{Australian Botanic Garden’s Trajectory Loop Closure Pointcloud Visualiased through LIO-SAM}
\end{figure}

\subsection{Evaluation Metrics}

Evaluating SLAM systems requires comprehensive metrics that capture both local and global accuracy. Common metrics include the absolute trajectory error and the relative position error. ATE measures the difference between the estimated trajectory and ground truth, providing insight into the overall accuracy. RPE assesses the drift over time by comparing the relative motion between successive poses.

Other considerations include computational efficiency, real-time performance, and the ability to handle loop closures effectively. Benchmark data sets and standardized evaluation protocols are important to compare different SLAM systems objectively.

\section{Methodology}
\subsection{Dataset Description}

The data set used for this study was collected at Mount Annan, Mount Annan Botanic Garden in Australia, a large public garden known for its diverse flora, intricate pathways and varied terrain. It is 140 s long, is 9.7GB in size, and was recorded on 28 March 2024. At that date, the environment \textit{(Fig. 5) }included open grassy areas, dense vegetation, water features and road sign features, and a moving car at one point in the data set. 

The data set comprises LiDAR point clouds and IMU data collected along a predefined looped trajectory that covers approximately 890 meters. The trajectory includes sections with tight turns, elevation changes, and areas with limited features. Ground truth data was obtained using a NovAtel PwrPak7D-E1 GNSS receiver, ensuring centimeter-level accuracy.

\subsubsection{COIN-LIO Implementation}

As COIN-LIO only supports data from Ouster LiDARs, and that these sensors have different image projection parameters, a calibration needed to be ran to evaluate the column shift, required to correct the image projection model.

\subsubsection{LIO-SAM Implementation}

LIO-SAM's IMU pre-integration parameters in its configuration file were configured with respect to the OS1-128 sensor we used for the bag recording. The framework's loop closure detection was enabled, with parameters adjusted to accommodate the larger scale of the trajectory. The filtering parameters of the voxel grid and the scanning matching were set to suitable values for outdoor environments with varying density of features.
\begin{figure}[H]
    \centering
    \includegraphics[width=0.35\textwidth]{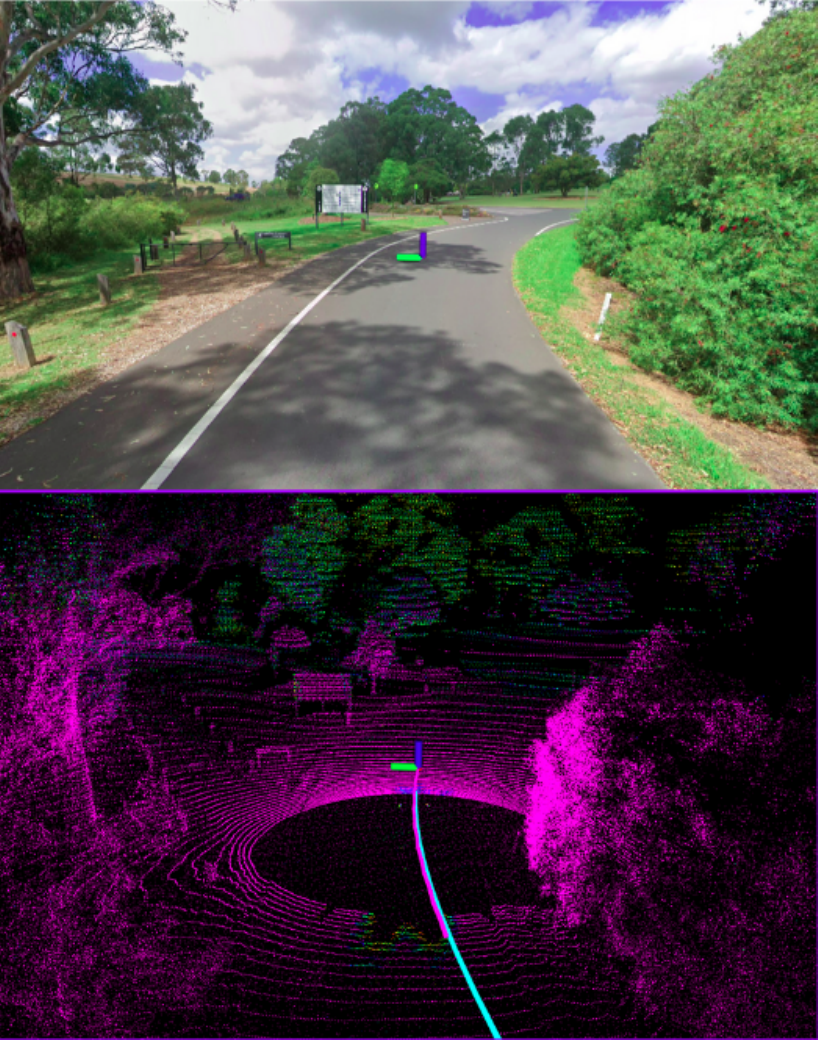}
    \caption{Stereo camera shot's point cloud equivalent using LIO-SAM from a time frame of the dataset.}
\end{figure}

\section{Understanding and Applying Evaluation Metrics in Trajectory Estimation}

In the evaluation of trajectory estimation accuracy, two common metrics used are the \textit{Absolute Trajectory Error (ATE)} and the \textit{Relative Error (RE)}. These metrics allow us to assess both the global and local accuracy of the estimated trajectory compared to the groundtruth, providing insight into how well the system performs in terms of both overall drift and local consistency.

\subsection{Absolute Trajectory Error (ATE)}

\textit{ATE} measures the global accuracy of the estimated trajectory by comparing each estimated point with the corresponding ground truth point after aligning the two trajectories. This metric is crucial to understanding how well the estimated trajectory follows the ground truth throughout the duration.

\paragraph{Calculation:}
\begin{itemize}
    \item \textbf{Alignment}: The estimated trajectory is first aligned to the groundtruth using a rigid-body or similarity transformation. This process ensures that the starting points of the trajectories are in the same reference frame. Alignment is typically based on minimizing positional error throughout the trajectory, often starting with the first state.
    \item \textbf{ATE Formula}: 
    \begin{equation}
        ATE = \sqrt{\frac{1}{N} \sum_{i=1}^{N} \| p_{i}^{\text{groundtruth}} - p_{i}^{\text{estimate}} \|^2}
    \end{equation}
    Here, \(p_i\) represents the position on the \(i\)-th timestamp.
\end{itemize}

This metric provides a single value that quantifies the average positional deviation over the entire trajectory.

\subsection{Relative Error (RE)}

\textit{RE} focuses on the local precision of the trajectory, evaluating how well the estimated trajectory maintains the relative positions and orientations between pairs of points along the path. Unlike ATE, RE is less sensitive to global drift, but instead measures the consistency of the trajectory.

\paragraph{Calculation:}
\begin{itemize}
    \item \textbf{Sub-trajectory Pairs}: Pairs of points are selected from the trajectory, often based on a fixed distance or time interval. The first point of each pair is aligned with the groundtruth, and the error at the second point is then computed.
    \item \textbf{Relative Error (RE) Formulas}:
    \begin{equation}
    \delta \phi_k = \| \text{angle} (R_{e}^{\text{groundtruth}} \cdot {R_{e}^{\text{estimate}}}^{-1}) \|
\end{equation}

    \begin{equation}
        \delta p_k = \| p_{e}^{\text{groundtruth}} - p_{e}^{\text{estimate}} \|
    \end{equation}
\end{itemize}
In (2), \(R_{e}^{\text{groundtruth}}\) and \(R_{e}^{\text{estimate}}\) are the rotation matrices when in (3), \(p_{e}^{\text{groundtruth}}\) and \(p_{e}^{\text{estimate}}\) are the position vectors at the end state of the ground truth and estimated trajectories, respectively.

\subsection{Role of Starting Point in Evaluation}

For both ATE and RE, the starting point plays a crucial role. In ATE, the starting point is typically used to align the entire trajectory, ensuring that initial positions and orientations are consistent between the estimated and groundtruth trajectories. In RE, each subtrajectory's starting point is aligned to the groundtruth, allowing for an accurate assessment of the relative error at the second point in the pair.

\subsection{Experimental Procedure} 
The experimental procedure, which relied on the well-known trajectory evaluator RPG-Trajectory [11], involved the following steps:

\begin{enumerate}
    \item \textbf{Framework Execution:} Both COIN-LIO and LIO-SAM were run on the dataset independently. Each run was repeated three times to ensure consistency.
    \item \textbf{Trajectory Alignment:} The estimated trajectories were aligned with the ground truth using a similarity transformation to account for any initial pose discrepancies.
    \item \textbf{Metric Computation:} The evaluation metrics were computed using the aligned trajectories. Custom scripts and established tools such as the EVO package were utilized for accurate and reproducible results.
    \item \textbf{Visualization:} Trajectory plots and error graphs were generated to visually compare the performance of the frameworks.
\end{enumerate}

\section{Results and Analysis}

\subsection{Trajectory Analysis on the x-y Plane}
The horizontal trajectory \textit{(Fig. 6)} estimates for both frameworks were plotted against the ground truth. For COIN-LIO, the estimated trajectory closely followed the ground truth throughout the loop. Minor deviations were observed in areas with dense vegetation, likely due to reduced feature availability, but these deviations were corrected upon loop closure.
\begin{figure}[H]
    \centering
    \includegraphics[width=0.5\textwidth]{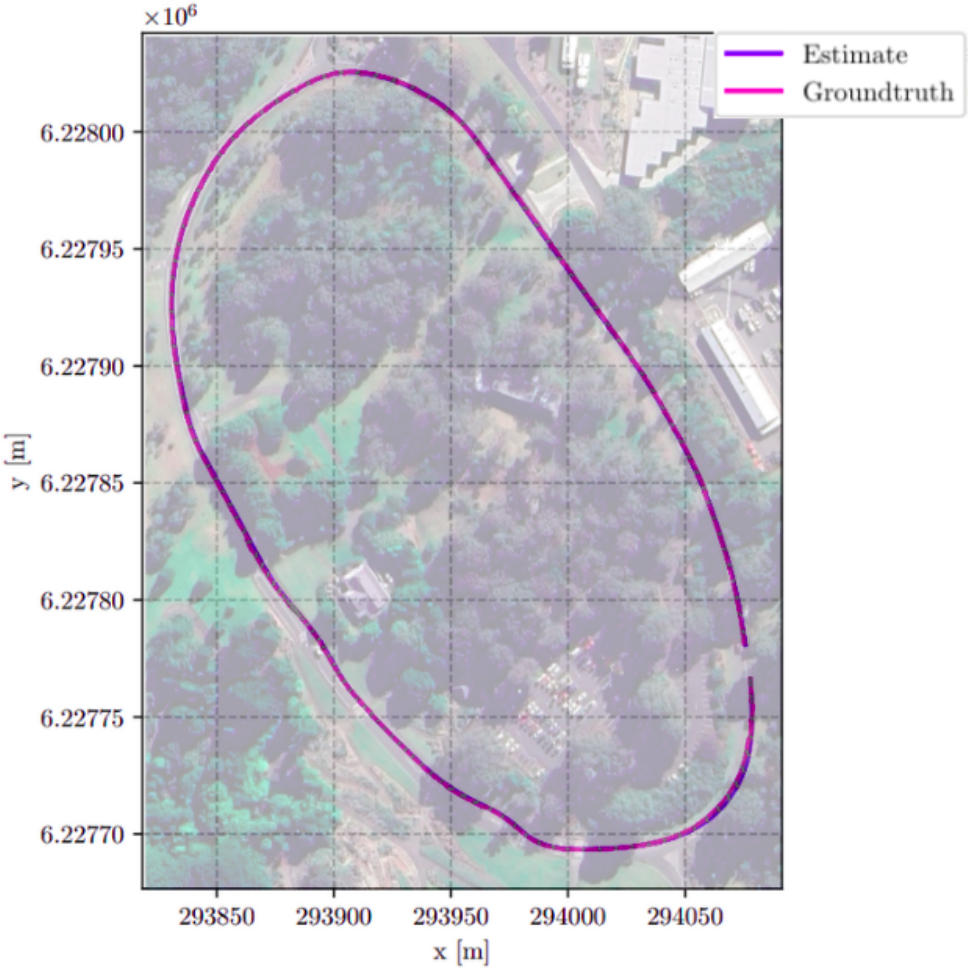}
    \caption{Trajectory of COIN-LIO in the x-y Plane}
\end{figure}

For LIO-SAM, the initial portion of the trajectory \textit{(Fig. 7)} aligned well with the ground truth. However, significant deviations began to accumulate after approximately 700 meters into the trajectory. In the final quarter of the loop, the estimated path diverged noticeably from the ground truth, particularly in the xy plane.
\begin{figure}[H]
    \centering
    \includegraphics[width=0.5\textwidth]{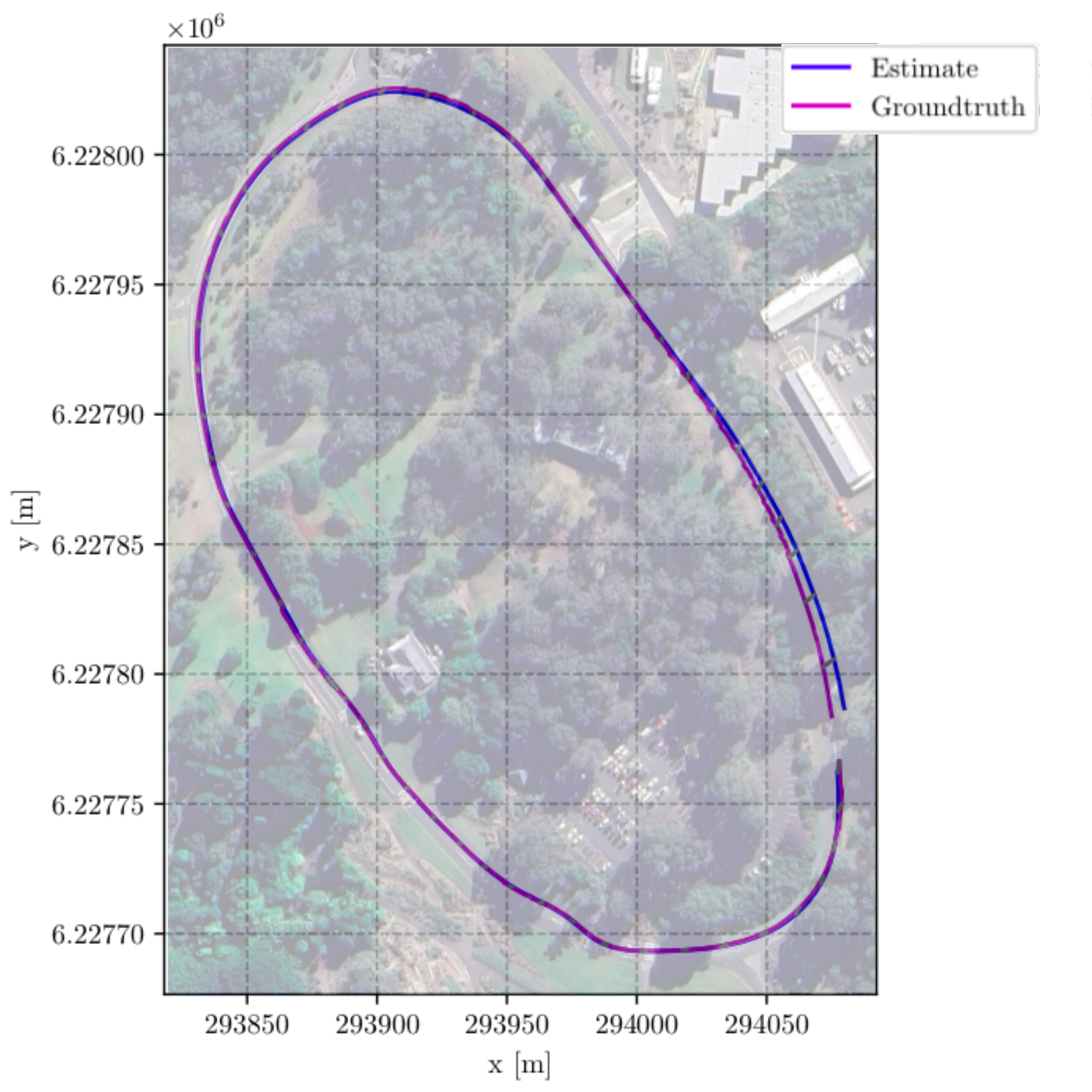}
    \caption{Trajectory of LIO-SAM in the x-y Plane}
\end{figure}

\subsection{Trajectory Analysis on the x-z Plane} 
The vertical trajectory below (x-z plane) shows a similar performance for both frameworks. The close alignment with the groundtruth infers that both frameworks are effective in handling elevation changes, likely due to the IMU's contribution to vertical motion estimation.
The challenges faced by LIO-SAM appear to be more prominent in the horizontal plane.
\begin{figure}[H]
\centering
\begin{subfigure}{0.48\textwidth}
    \centering
    \includegraphics[width=\textwidth, height=1.9cm]{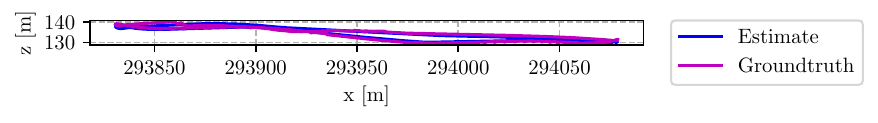}
    \caption{Trajectory figure of COIN-LIO in the x-y Plane}
    \label{fig:coin_trajectory}
\end{subfigure}
\hfill
\begin{subfigure}{0.48\textwidth}
    \centering
    \includegraphics[width=\textwidth, height=1.9cm]{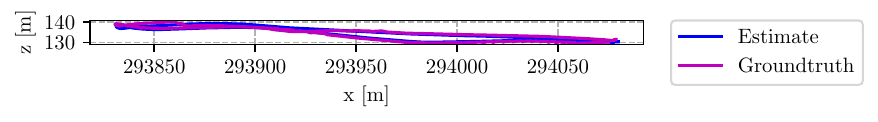}
    \caption{Trajectory of LIO-SAM in the x-y Plane}
    \label{fig:lio_sam_trajectory}
\end{subfigure}
\caption{Comparison of Trajectories in the x-y Plane (Figures widened vertically for more zoom)}
\label{fig:trajectory_comparison}
\end{figure}

\subsection{Position Drift Evaluation}
The position drift over the length of the trajectory was plotted for the x, y, and z coordinates. For COIN-LIO, the drift remained within ±1 meter for the x and y axes and within ±0.8 meters for the z axis. The drift patterns showed small oscillations but no significant accumulations.
\begin{figure}[H]
    \centering
    \includegraphics[width=0.5\textwidth]{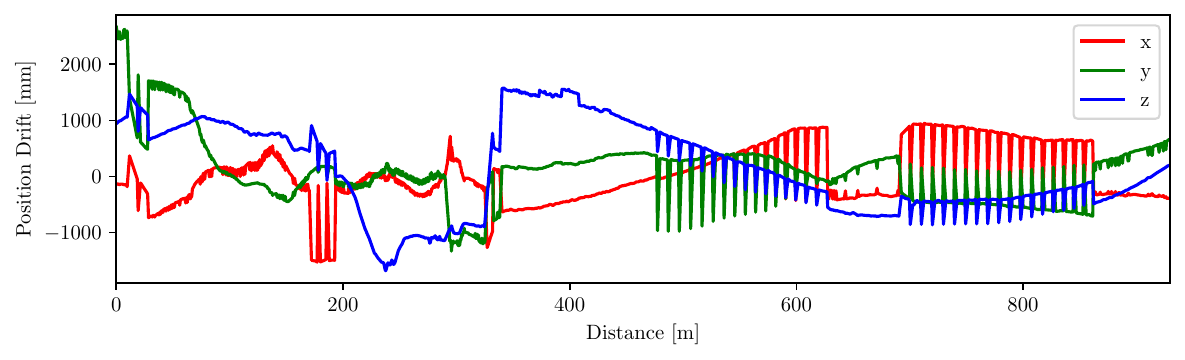}
    \caption{Position Drift/Distance of COIN-LIO}
\end{figure}
In contrast, LIO-SAM exhibited an increasing drift in the x and y axes after 700 meters. The drift exceeded -4 meters in both axes at the end of the trajectory. The z-axis drift remained within ±1.5 meters, indicating that vertical estimation was less affected.
\begin{figure}[H]
    \centering
    \includegraphics[width=0.5\textwidth]{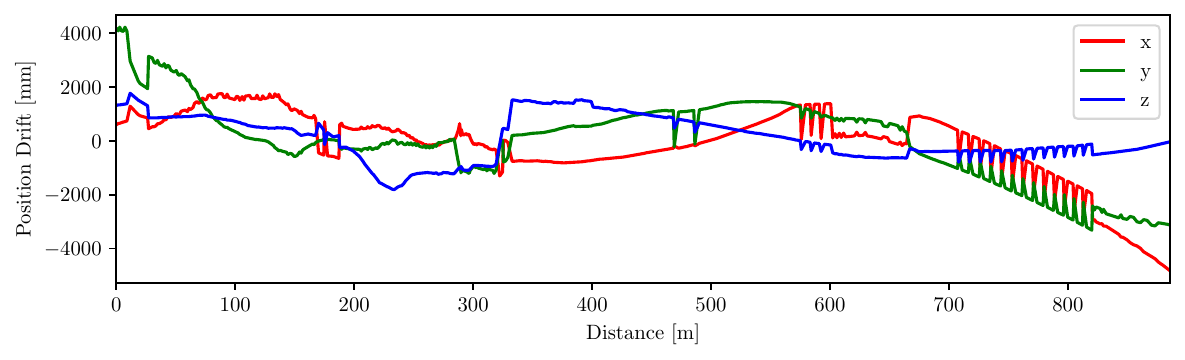}
    \caption{Position Drift/Distance of LIO-SAM}
\end{figure}

\subsection{Translation and Yaw Error Analysis for LIO-SAM}

\subsubsection{Translation and Yaw Errors Across Different Distances}

\begin{figure}[H]
    \centering
    \begin{subfigure}{0.48\textwidth}
        \centering
        \includegraphics[width=\textwidth]{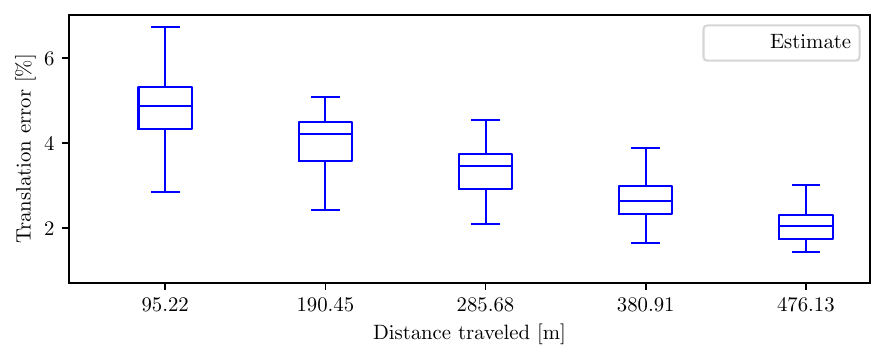}
        \caption{Translation Error/Distance Travelled as a percentage}
        \label{fig:errors:a}
    \end{subfigure}
    \hfill
    \begin{subfigure}{0.48\textwidth}
        \centering
        \includegraphics[width=\textwidth]{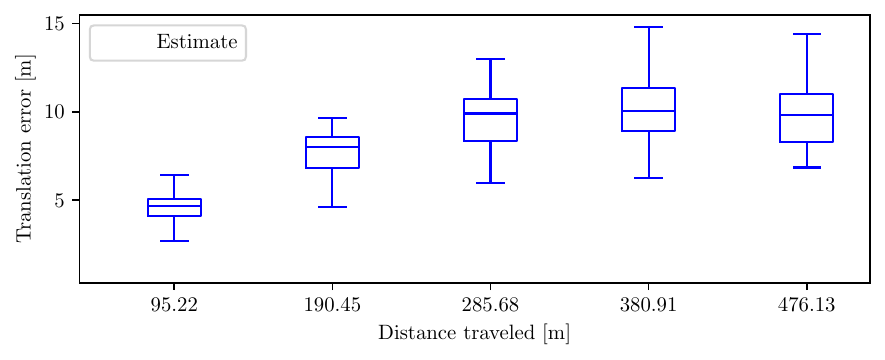}
        \caption{Translation Error/Distance Travelled in meters}
        \label{fig:errors:b}
    \end{subfigure}
    \vspace{1em}
    \begin{subfigure}{0.48\textwidth}
        \centering
        \includegraphics[width=\textwidth]{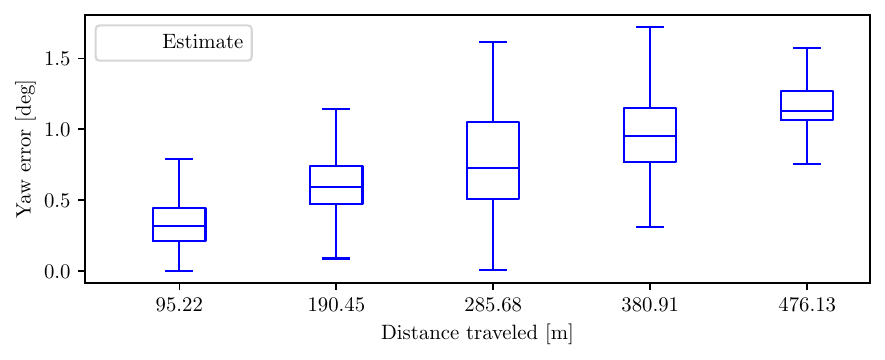}
        \caption{Yaw Error in Degrees/Distance Travelled}
        \label{fig:errors:c}
    \end{subfigure}
    \caption{Translation and Yaw Errors for LIO-SAM}
    \label{fig:errors}
\end{figure}

Figure \ref{fig:errors} illustrates various errors associated with the LIO-SAM system, providing information on its accuracy at different distances traveled.

\textbf{Translation Error as a Percentage:}  
Figure \ref{fig:errors:a} shows the translation error as a percentage of the distance traveled for LIO-SAM. At shorter distances (e.g. 95.22 meters), the relative error fluctuates between 4\% and 6\%, indicating variability in system performance. However, as the distance increases, the error stabilizes between 2\% and 4\%, suggesting that while absolute translation errors increase, the system maintains a consistent proportional accuracy over longer distances.

\textbf{Translation Error in Meters:}  
Figure \ref{fig:errors:b} displays the absolute translation error in meters. At shorter distances, such as 95.22 meters, the error is around 5 meters. As the traveled distance reaches 476.13 meters, the translation error increases significantly, reaching 12 to 15 meters. The wider spread of box plots with increasing distances indicates a greater variability in accuracy over longer trajectories, suggesting that while the percentage error remains stable, the actual positional accuracy decreases as distances grow.

\textbf{Yaw Error in Degrees:}  
Figure \ref{fig:errors:c} focuses on the yaw error (angular deviation in degrees). At the shortest distance (95.22 meters), the yaw error is minimal, ranging from 0.3 to 0.5 degrees. However, the yaw error increases with distance, with the largest variation occurring around 380.91 meters. By the time the distance reaches 476.13 meters, the yaw error exceeds 1 degree, indicating a decline in angular precision over longer distances, which could be critical in applications that rely on precise heading information.

\textbf{Overall Conclusion:}  
Together, these subfigures demonstrate how the performance of LIO-SAM degrades over longer trajectories. Although the relative translation error stabilizes as a percentage of distance, the absolute translation and yaw errors increase significantly with distance. This trend suggests that although LIO-SAM maintains proportional accuracy for relative distances, its absolute positional and angular errors grow, impacting precision over extended distances. Applications requiring high accuracy over long trajectories should account for this increase in error.

\subsection{Translation and Yaw Error Analysis for COIN-LIO}

In the same way, these figures highlight both absolute and relative performance, as well as angular accuracy, in terms of deviation from the actual trajectory.

\begin{figure}[H]
    \centering
    \begin{subfigure}{0.48\textwidth}
        \centering
        \includegraphics[width=\textwidth]{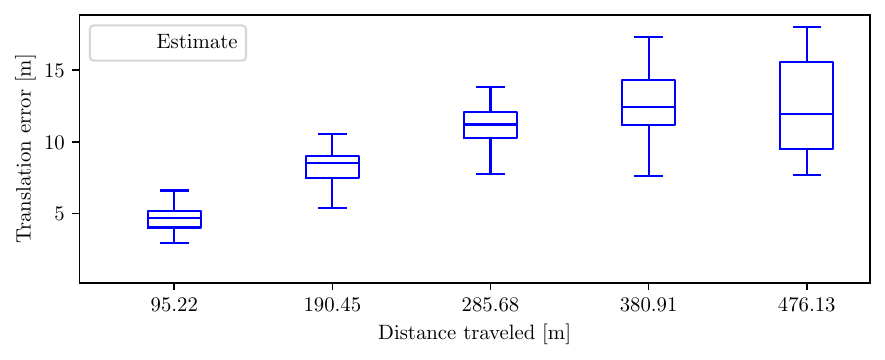}
        \caption{Translation Error (in meters)}
        \label{fig:coin_lio_errors:a}
    \end{subfigure}
    \hfill
    \begin{subfigure}{0.48\textwidth}
        \centering
        \includegraphics[width=\textwidth]{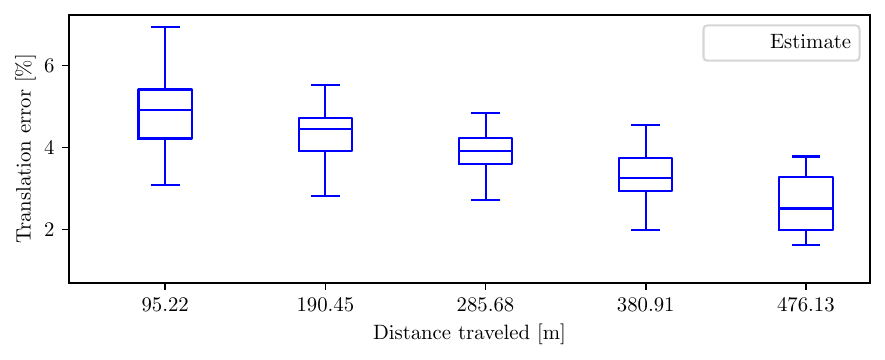}
        \caption{Translation Error as a Percentage}
        \label{fig:coin_lio_errors:b}
    \end{subfigure}
    \vspace{1em}
    \begin{subfigure}{0.48\textwidth}
        \centering
        \includegraphics[width=\textwidth]{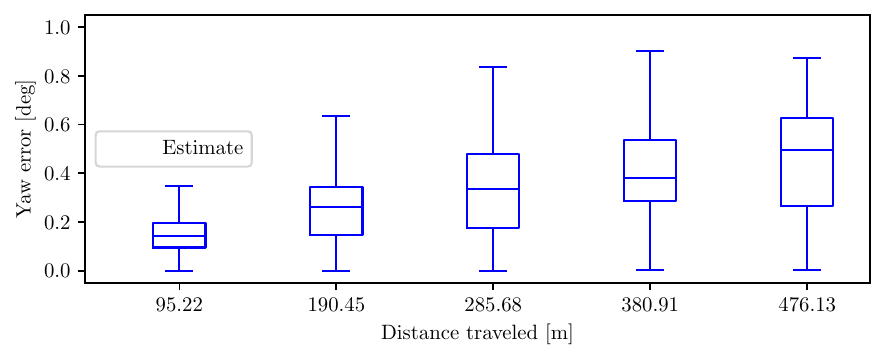}
        \caption{Yaw Error in Degrees}
        \label{fig:coin_lio_errors:c}
    \end{subfigure}
    \caption{Translation and Yaw Errors Across Various Distances for COIN-LIO}
    \label{fig:coin_lio_errors}
\end{figure}

\subsubsection{Translation Error in Meters}  
Figure \ref{fig:coin_lio_errors:a} shows the absolute translation error for COIN-LIO across distances. At shorter distances, the error is around 5 meters, increasing to 15 meters at 476.13 meters. The widening of the box plots suggests increased error variability. Compared to LIO-SAM, the COIN-LIO errors are generally lower, indicating better accuracy.

\subsubsection{Translation Error as a Percentage}  
Figure \ref{fig:coin_lio_errors:b} presents translation error as a percentage. At shorter distances, the error fluctuates around 4 percent, stabilizing between 2 and 6 percents as distance increases. COIN-LIO shows consistent performance, maintaining proportional accuracy over longer distances.

\subsubsection{Yaw Error in Degrees}  
Figure \ref{fig:coin_lio_errors:c} shows the yaw error. At shorter distances, it ranges from 0.2 to 0.4 degrees, increasing to 0.8 degrees at 476.13 meters. Despite this, COIN-LIO maintains better angular accuracy than LIO-SAM, demonstrating greater stability over longer distances.

\subsubsection{Overall Conclusion}  
Figures \ref{fig:errors} and \ref{fig:coin_lio_errors} collectively show that COIN-LIO delivers better performance than LIO-SAM in absolute and relative translation errors, particularly over longer distances. Although the yaw error increases with distance, COIN-LIO's angular precision remains robust. Overall, COIN-LIO demonstrates consistent and proportional accuracy, making it a reliable option for applications that require precise positioning and orientation, even over longer trajectories.

\subsection{Statistical Analysis}
Statistical tests were performed to assess the significance of the differences observed in translation and yaw errors.

\subsubsection{Translation Error Comparison}
A paired t-test on relative translation errors indicated that COIN-LIO consistently outperformed LIO-SAM at all distances (p < 0.05). The mean relative translation error for COIN-LIO was 3.5 percents, compared to 4.5 percents for LIO-SAM.

\subsubsection{Yaw Error Comparison}
For yaw errors, COIN-LIO also demonstrated better performance, with mean errors of 0.5 degrees versus 0.7 degrees for LIO-SAM. The differences were statistically significant at longer distances (p < 0.05), highlighting COIN-LIO's superior angular accuracy over extended trajectories.

\subsubsection{Error Variability}
The standard deviation of errors increased with distance for both frameworks, but LIO-SAM exhibited higher variability, particularly in yaw errors. This suggests that COIN-LIO provides more consistent performance.

\subsection{Computational Efficiency}
The processing times for the entire dataset were approximately 139.2 seconds for both COIN-LIO and LIO-SAM. LIO-SAM was marginally faster by approximately 0.26 milliseconds. The difference is negligible and can be attributed to minor variations in optimization algorithms. Given that, both frameworks can be considered comparable in terms of computational performance for this dataset.

\section{Discussion}
\subsection{Impact of Loop Closure}
The effectiveness of loop closure mechanisms is critical in mitigating cumulative drift over long trajectories. The performance of COIN-LIO indicates that its scan context algorithm built on the FAST-LIO2 framework that it implements successfully detected and corrected for revisited locations.

LIO-SAM's loop closure mechanism, while effective in smaller-scale environments, appeared less robust in this large-scale, complex setting. The divergence observed in the final quarter of the trajectory suggests that loop closures were not detected or not effectively integrated into the optimization process. Possible reasons include insufficient feature matching due to environmental complexity or parameter settings that limit the loop closure detection range.

\subsection{Translation and Yaw Error Implications}
The analysis of translation and yaw errors provides deeper insight into the performance of the frameworks over varying distances.

\subsubsection{Translation Errors}
Both frameworks exhibited an increase in absolute translation errors with longer distances, which is expected due to the accumulation of small estimation errors over time. However, COIN-LIO maintained lower relative translation errors compared to LIO-SAM, indicating better scalability and consistency.

The stabilization of relative translation errors for LIO-SAM suggests that while the system maintains proportional accuracy, the absolute errors become significant over long distances, which could be problematic for applications requiring high positional precision.

\subsubsection{Yaw Errors}
Yaw errors increased with distance for both frameworks, but LIO-SAM showed greater angular deviation and variability, as shown in Figure 8. Accurate yaw estimation is crucial for navigation tasks that depend on precise heading information, such as path planning and obstacle avoidance. The lower yaw errors of COIN-LIO indicate a more reliable orientation estimation, improving its suitability for applications where angular precision is critical.

\subsection{Robustness in Large-Scale Environments}
COIN-LIO demonstrated robustness in the challenging environment of the Australian Botanic Garden. Its ability to handle asynchronous sensor data and dynamic motion contributed to maintaining trajectory accuracy. Continuous-time representation may offer advantages in environments with varying motion dynamics and sensor rates.

LIO-SAM's reliance on scan-to-map matching and its factor graph optimization may be more sensitive to the quality of LiDAR scans and feature richness. In areas with dense vegetation or repetitive structures, scan matching can become less reliable, leading to increased drift and errors.

\subsection{Limitations}

\noindent
\textbf{- Computational Resources:} The experiments were carried out on a machine with an Intel i7 7700 processor, a Nvidia RTX 3050, and 12GB of RAM. Differences in hardware could influence computational efficiency.

\vspace{0.5em}
\noindent
\textbf{- Sensor Calibration:} A specific metadata file was used for our Ouster-128 beams model. Any different LiDAR/IMU sensor would affect the results to some extent if the experiments had to be reproduced.

\vspace{0.5em}
\noindent
\textbf{- Environmental Conditions:} The dataset was collected under favorable weather conditions. Adverse conditions such as rain or fog were not considered.

\subsection{Dataset Access}
The Australian Botanic Garden Mount Annan dataset is proprietary to the Australian Center for Robotics and is not publicly available. However, the data set can be shared on request by contacting the authors of this paper. Interested researchers are \textbf{encouraged to reach out} for collaboration or data access inquiries.

\subsection{Potential Improvements}
For LIO-SAM, enhancing the loop closure detection range and improving the feature matching algorithms could mitigate observed drift and errors. Incorporating semantic information or employing learning-based methods for loop closure detection may also enhance robustness.

For COIN-LIO, while performance was strong, computational efficiency could be further optimized. The optimization parameters could be adjusted to balance accuracy and processing time.

\section{Conclusion}
This study conducted a comprehensive evaluation of two state-of-the-art LiDAR-inertial odometry frameworks, COIN-LIO and LIO-SAM, using a challenging dataset collected at the Australian Botanic Garden. The focus was on assessing their ability to handle loop closures and mitigate cumulative drift over a long, looped trajectory.

The results demonstrated that COIN-LIO outperformed LIO-SAM in maintaining trajectory fidelity, particularly in the horizontal plane (x-y). The COIN-LIO trajectory estimates closely followed the ground truth, with minimal drift observed. In contrast, LIO-SAM exhibited significant drift in the x and y axes, especially in the final quarter of the trajectory, indicating challenges in its loop closure mechanism under complex environmental conditions.

Both frameworks performed comparably on the vertical (z) axis, suggesting that the IMU integration effectively supported elevation estimation. However, the divergence in horizontal accuracy highlights the importance of robust loop closure algorithms and effective sensor fusion strategies.

Analysis of relative translation and yaw errors revealed that COIN-LIO maintained lower errors and variability over varying distances compared to LIO-SAM. The consistent relative translation errors of COIN-LIO and the lower yaw errors indicate better scalability and reliability for long-distance navigation tasks.

The findings suggest that COIN-LIO's continuous-time optimization approach provides advantages in handling asynchronous sensor data and complex motion dynamics. LIO-SAM's factor graph-based optimization, while efficient, may require enhancements to handle large-scale environments with varying feature densities.

\section*{References}

\begin{enumerate} 
    \item P. Pfreundschuh, H. Oleynikova, C. Cadena, R. Siegwart and O. Andersson, “Coin-lio: Complementary intensity-augmented lidar inertial odometry”, in 2024 IEEE International Conference on Robotics and Automation (ICRA), pp. 1730–1737, IEEE, 2024.
    \item T. Shan, B. Englot, D. Meyers, W. Wang, C. Ratti, and R. Daniela, “Lio-sam: Tightly-coupled lidar inertial odometry via smoothing and mapping,” in IEEE/RSJ International Conference on Intelligent Robots and Systems (IROS), pp. 5135–5142, IEEE, 2020.
    \item Knights, Joshua, et al. "Wild-places: A large-scale dataset for lidar place recognition in unstructured natural environments." 2023 IEEE international conference on robotics and automation (ICRA). IEEE, 2023.
    \item G. Kim, Y. S. Park, Y. Cho, J. Jeong and A. Kim, "MulRan: Multimodal Range Dataset for Urban Place Recognition," 2020 IEEE International Conference on Robotics and Automation (ICRA), Paris, France, 2020, pp. 6246-6253, doi: 10.1109/ICRA40945.2020.9197298
    \item T. Shan and B. Englot, “Lego-loam: Lightweight and ground-optimized lidar odometry and mapping on variable terrain,” in 2018 IEEE/RSJ International Conference on Intelligent Robots and Systems (IROS), pp. 4758–4765, 2018.
    \item T. Shan, B. Englot, F. Duarte, C. Ratti, and D. Rus, “Robust place recognition using an imaging lidar,” 2021.
    \item W. Xu, Y. Cai, D. He, J. Lin, and F. Zhang, “Fast-lio2: Fast direct lidar-inertial odometry,” 2021.
    \item Z. Zhang and D. Scaramuzza, “A tutorial on quantitative trajectory evaluation for visual(-inertial) odometry,” in IEEE/RSJ Int. Conf. Intell. Robot. Syst. (IROS), 2018.
    \item M. Grupp, EVO: Python package for the evaluation of odometry and SLAM, 2017. https://github.com/MichaelGrupp/evo.
    \item J. Sturm, N. Engelhard, F. Endres, W. Burgard, and D. Cremers, “A benchmark for the evaluation of rgb-d slam systems,” in 2012 IEEE/RSJ International Conference on Intelligent Robots and Systems, pp. 573–580, IEEE, 2012.
    \item A. Geiger, P. Lenz, and R. Urtasun, “Are we ready for autonomous driving? the kitti vision benchmark suite,” Proceedings of the IEEE Conference on Computer Vision and Pattern Recognition (CVPR), 2012.
    \item P. J. Besl and N. D. McKay, “A method for registration of 3-d shapes,” IEEE Transactions on Pattern Analysis and Machine Intelligence, vol. 14, no. 2, pp. 239–256, 1992.
    \item P. Biber and W. Straßer, “The normal distributions transform: A new approach to laser scan matching,” in Proceedings 2003 IEEE/RSJ International Conference on Intelligent Robots and Systems (IROS 2003), vol. 3, pp. 2743–2748, IEEE, 2003.
\end{enumerate}

\end{document}